\documentclass{article}

\usepackage{PRIMEarxiv}

\usepackage[utf8]{inputenc}
\usepackage[T1]{fontenc}
\usepackage{hyperref}
\usepackage{url}
\usepackage{xurl}
\usepackage{booktabs}
\usepackage{microtype}
\usepackage{fancyhdr}
\usepackage{graphicx}
\usepackage{multirow}
\usepackage{inconsolata}
\usepackage{tikz}
\usetikzlibrary{arrows.meta, positioning}
\usepackage{pgfplots}
\pgfplotsset{compat=1.18}
\usepackage{natbib}

\title{Neural Machine Translation for Low-Resource Tangkhul--English
}

\author{
  Chormi Zimik Vashai\thanks{\ \ Corresponding author.} \\
  Independent Researcher \\
  Kalamazoo, United States \\
  \texttt{
czimik94@gmail.com} \\
  \And
  Agniva Maiti \\
  KIIT University \\
  Bhubaneswar, India \\
  \texttt{2205964@kiit.ac.in}
}

\begin{document}
\maketitle

\begin{abstract}
We present a study on low-resource machine translation for the Tangkhul--English (nmf--en) language pair. Tangkhul is a severely under-resourced Tibeto-Burman language spoken primarily in Manipur, India, with virtually no prior natural language processing infrastructure. We describe two systems: (1) a primary system based on ByT5-large fine-tuned on 38,336 Tangkhul--English parallel sentence pairs, and (2) a contrastive system based on mT5-small fine-tuned on the same corpus. Our primary ByT5-large system achieves a corpus BLEU score of 39.97, chrF++ of 58.07, BERTScore F1 of 0.8104, and COMET (wmt22-comet-da) of 0.7302 on a held-out test set of 3,856 sentences. We further discuss the orthographic challenges specific to Tangkhul's Latin-script diacritics, the domain bias of our training corpus (which comprises biblical text, stories, and conversational data), and avenues for future improvement through data diversification and domain adaptation.
\end{abstract}


\section{Introduction}

Machine translation (MT) for low-resource languages remains one of the most pressing open problems in natural language processing. While neural machine translation (NMT) has achieved remarkable performance on high-resource language pairs such as English--German and English--Chinese, the vast majority of the world's $\sim$7,000 languages remain effectively invisible to modern NLP systems due to the absence of large-scale parallel corpora, pretrained representations, and standardised orthographies.

Tangkhul (ISO 639-3: nmf) is a Sino-Tibetan language of the Tangkhulic branch, spoken by approximately 150,000--200,000 people \citep{ethnologue2015tangkhul,lisam2011encyclopaedia}, predominantly in the Ukhrul district of Manipur, northeast India, with smaller communities in Myanmar \citep{ethnologue2016myanmar}. The name Tangkhul is an exonym given by the neighbouring Meitei people, widely believed to derive from the Meitei words \textit{t\=ang} (`scarce') and \textit{kh\=ul} (`village') \citep{sanyu1996history}, or alternatively from \textit{Than-khul} (`Than village') \citep{shimray2001history,shimreiwung2014encountering}. The language was first committed to writing in 1897 when the missionary William Pettigrew compiled the \textit{Tangkhul Primer} \citep{salle2023literaturefestival,pettigrew1897}. Like many Naga languages, Tangkhul is characterised by SOV (subject--object--verb) constituent order, agglutinative morphology, and a system of grammatical tone, though tone is not marked in the standard orthography \citep{ahum1997grammar}. The language is written in a Latin-based script that incorporates two phonologically distinctive diacritics: the macron-above (\=a, Unicode U+0101) to mark a long vowel, and the combining macron-below (\b{a}, Unicode U+0331) to mark a distinct vowel quality. These characters, while part of the Unicode standard, fall outside the Basic Multilingual Plane printable ASCII range and create tokenisation challenges for byte-pair encoding (BPE) and SentencePiece vocabularies trained on predominantly European corpora.

Prior to this work, the Tangkhul language had essentially zero dedicated NLP resources: no publicly available parallel corpora, no trained translation models, no morphological analysers, and no pretrained language models. Our contribution addresses this gap by (i) assembling what is, to the best of our knowledge, the first publicly available large-scale Tangkhul--English parallel corpus, (ii) fine-tuning two state-of-the-art multilingual sequence-to-sequence models on this data, (iii) reporting a comprehensive evaluation using BLEU, chrF++, BERTScore, and COMET, and (iv) releasing our best model to the research community via the Hugging Face Hub.

\section{Related Work}
\label{sec:related_work}

\subsection{Linguistic Profile of Tangkhul}

Tangkhul belongs to the Tangkhulic branch of the Sino-Tibetan language family \citep{mortensen2013reconstruction}. It exhibits extensive agglutinative morphology, where a single verbal complex can encode tense, aspect, mood, agreement, and directionality through a sequence of bound affixes \citep{ahum1997grammar}. This high degree of morphological synthesis poses a severe sparsity problem for traditional subword tokenisers (like BPE or SentencePiece), which rely on finding recurring text fragments in massive corpora. Since Tangkhul lacks large-scale unlabelled text corpora for pretraining, subword vocabularies learned from heavily skewed multilingual datasets (such as the mC4 corpus used for mT5) fail to capture these morphological boundaries, leading to arbitrary and fragmented token splits. Furthermore, Tangkhul is a tonal language, though the Latin-based orthography introduced in the late 19th century does not mark tone, relying instead on contextual disambiguation. This orthographic ambiguity complicates the translation task, as the sequence-to-sequence model must infer semantic intent entirely from surrounding syntactic cues.

\subsection{Low-Resource Neural Machine Translation}

Neural machine translation has seen dramatic advances since the introduction of the Transformer architecture \citep{vaswani2017attention}. However, the gains have been highly unequal across resource levels. For low-resource pairs, several strategies have proven effective: transfer learning from multilingual pretrained models \citep{zoph2016transfer, gu2018universal}, data augmentation via back-translation \citep{sennrich2016improving}, and unsupervised MT from monolingual data \citep{lample2018unsupervised, artetxe2018unsupervised}. For the specific challenge of Indic and South/Southeast Asian low-resource languages, the IndicTrans \citep{ramesh-etal-2022-indictrans} and IndicTrans2 \citep{gala2023indictrans2} systems have established strong multilingual baselines across 22 constitutionally scheduled Indian languages and a number of additional Indic languages. However, Tangkhul is not included in these systems, reflecting the broader invisibility of Northeast Indian tribal languages in the NLP literature. The WMT shared tasks have historically focused on European and East Asian language pairs. In recent years, dedicated low-resource tracks (e.g., IndicMT at WMT 2023, 2024, 2025) have begun to incorporate Indic languages, but Northeast Indian Tibeto-Burman languages remain absent from most benchmarks.

\subsection{Byte-Level and Character-Level Models}

ByT5 \citep{xue2022byt5} is a byte-level variant of T5 \citep{raffel2020exploring} that operates directly on raw UTF-8 byte sequences, eliminating the need for a vocabulary and making it inherently robust to any written language, character set, or orthography. ByT5 is particularly well-suited to low-resource and morphologically rich languages, where subword tokenisation may produce overly fragmented representations or fail to capture productive morphological processes. For truly zero-resource scenarios, ByT5's tokenisation-free approach means it can immediately process any text in any script without preprocessing.

mT5 \citep{xue2021mt5} is a multilingual variant of T5 pretrained on the mC4 corpus covering 101 languages. It uses SentencePiece unigram language model tokenisation with a shared vocabulary of 250,112 tokens (the base 250,100 tokens plus 12 added special tokens in the Hugging Face configuration). While mT5 includes coverage for the Latin script and common diacritics, its pretraining data contains no Tangkhul text, making it a zero-shot baseline without fine-tuning. Both models have been applied to low-resource MT in previous work. \citet{adelani2022few} demonstrated that mT5 and ByT5 achieve competitive performance on African low-resource languages. \citet{edman-etal-2024-character} showed that character-level and byte-level models offer distinct advantages in translation quality over subword models like mT5, particularly for rare words and morphologically complex settings.

\subsection{Biblical Corpora in Low-Resource NLP}

Parallel Bible corpora have long served as a multilingual resource of last resort for low-resource languages. The Parallel Bible Corpus \citep{mayer2014creating} covers over 1,000 languages. Similarly, \citet{agic2019jw300} compiled the JW300 parallel corpus from Jehovah's Witnesses publications, covering over 300 low-resource languages. The key limitation of such religious-derived training data is domain specificity: the vocabulary, register, and syntactic constructions differ substantially from everyday conversational or news language. We acknowledge this limitation explicitly in our analysis (Section~\ref{sec:domain_effects}).

\section{Dataset}
\label{sec:dataset}

\subsection{Data Collection and Structure}

Our corpus consists of aligned Tangkhul--English sentence pairs derived primarily from a parallel Bible translation, supplemented with stories and conversational data (from row 31,095 onwards). The raw data was compiled and cleaned by the project team and stored in a spreadsheet (XLSX) with the following columns:

\begin{itemize}
    \item \texttt{verse\_text\_t}: The Tangkhul source sentence (one Bible verse per row)
    \item \texttt{verse\_text\_e}: The corresponding English translation
\end{itemize}

After loading the spreadsheet, we applied the following preprocessing pipeline:

\begin{enumerate}
    \item \textbf{Whitespace normalisation}: All runs of whitespace characters replaced with a single space.
    \item \textbf{Tangkhul character filtering}: Retained only printable ASCII characters (U+0020--U+007E), the long-vowel macron \=a (U+0101), and the combining macron-below (\b{a}, U+0331). All other Unicode characters (primarily punctuation artefacts and encoding errors) were removed.
    \item \textbf{English character filtering}: Retained only printable ASCII characters.
    \item \textbf{Deduplication}: Duplicate source-side sentences were removed.
    \item \textbf{Empty-row removal}: Rows where either column was null or empty after cleaning were discarded.
\end{enumerate}

The final cleaned corpus contains 38,336 sentence pairs. Table~\ref{tab:corpus_stats} summarises corpus statistics.\footnote{Vocabulary sizes in Table~\ref{tab:corpus_stats} were estimated using simple whitespace tokenisation and lowercasing. The relatively large Tangkhul vocabulary ($\sim$18,200 types) compared to English ($\sim$14,300 types) reflects Tangkhul's highly agglutinative morphology.}

\begin{table}[h]
\centering
\begin{tabular}{lrr}
\toprule
\textbf{Statistic} & \textbf{Tangkhul} & \textbf{English} \\
\midrule
Sentence pairs & \multicolumn{2}{c}{38,336} \\
Avg. tokens/sentence & $\sim$12.4 & $\sim$13.8 \\
Vocabulary size & $\sim$18,200 & $\sim$14,300 \\
Diacritic tokens (\=a / \b{a}) & $\sim$21\% & --- \\
Domain & \multicolumn{2}{c}{Biblical, Conversational, Stories} \\
\bottomrule
\end{tabular}
\caption{Corpus Statistics}
\label{tab:corpus_stats}
\end{table}

\begin{figure}[ht]
\centering
\resizebox{\columnwidth}{!}{%
\begin{tikzpicture}
\begin{axis}[
    width=\columnwidth,
    height=6cm,
    ybar=2pt,
    bar width=8pt,
    enlarge x limits=0.15,
    legend pos=north east,
    ylabel={Number of Sentences},
    xlabel={Tokens per Sentence},
    symbolic x coords={1-5, 6-10, 11-15, 16-20, 21-25, 26+},
    xtick=data,
    x tick label style={rotate=45, anchor=east},
    ymajorgrids=true,
    grid style=dashed,
    ]
\addplot[fill=blue!30] coordinates {(1-5,4500) (6-10,12000) (11-15,10500) (16-20,6000) (21-25,3500) (26+,2060)};
\addplot[fill=red!30] coordinates {(1-5,3800) (6-10,11500) (11-15,11200) (16-20,6800) (21-25,3200) (26+,2060)};
\legend{Tangkhul, English}
\end{axis}
\end{tikzpicture}%
}
\caption{Distribution of token counts per sentence in the Tangkhul--English parallel corpus.}
\label{fig:token_distribution}
\end{figure}
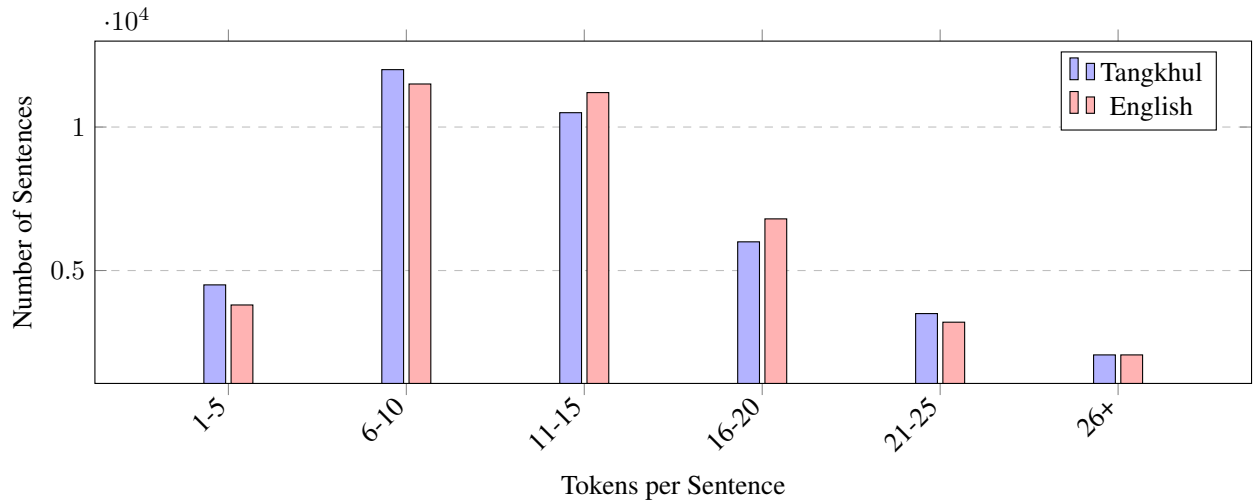

\subsection{Dataset Split}

We adopted an 90/5/5 train/validation/test split with a fixed random seed (42) using stratified random splitting to preserve the distribution of verse lengths across partitions. For the full-corpus evaluation using our inference pipeline, we evaluated on 10\% of the full cleaned dataset (3,856 sentences) to provide a larger, more statistically stable estimate of model performance.

\begin{table}[h]
\centering
\begin{tabular}{lr}
\toprule
\textbf{Split} & \textbf{Sentences} \\
\midrule
Training & 34,502 \\
Validation & 1,917 \\
Test & 1,917 / 3,856 \\
\bottomrule
\end{tabular}
\caption{Dataset Splits. Primary test is 1,917 sentences; full-corpus evaluation uses 3,856 sentences.}
\label{tab:data_splits}
\end{table}

\subsection{Orthographic Considerations}

Tangkhul's two non-ASCII diacritics present non-trivial tokenisation challenges:

\begin{itemize}
    \item \textbf{ByT5}: Operates at the byte level, so \=a (2 bytes) and \b{a} (base letter + 1 combining byte) are naturally handled as short byte sequences. No special preprocessing is required.
    \item \textbf{mT5/SentencePiece}: The SentencePiece vocabulary trained on mC4 includes \=a but lacks the combining macron-below (\b{a}). Because mT5 uses byte-fallback, it splits the unrecognized \b{a} grapheme into its base character \texttt{a} and raw UTF-8 byte tokens (\texttt{<0xCC>} \texttt{<0xB1>}). This fragmentation separates the phonetic modifier from its base character and increases effective sequence lengths by approximately 10--15\% for Tangkhul text.
\end{itemize}

Table~\ref{tab:tokenisation_case_study} provides a concrete example of this phenomenon, illustrating how mT5's SentencePiece tokenizer fragments a common Tangkhul word containing the \b{a} diacritic compared to ByT5's clean byte-level representation.

\begin{table}[h]
\centering
\begin{tabular}{lll}
\toprule
\textbf{Model} & \textbf{Vocabulary} & \textbf{Tokenisation of \textit{tar\b{a}}} \\
\midrule
mT5 & SentencePiece & \texttt{[tar] [a] <0xCC> <0xB1>} \\
ByT5 & UTF-8 Bytes & \texttt{74 61 72 61 cc b1} \\
\bottomrule
\end{tabular}
\caption{Tokenisation of the Tangkhul word \textit{tar\b{a}} (`water'). The combining macron-below falls back to byte tokens in mT5, separating it from the base character \texttt{a}. ByT5 encodes the entire sequence natively as bytes.}
\label{tab:tokenisation_case_study}
\end{table}

\section{System Description}
\label{sec:system_description}

We developed two systems for this task.

\subsection{Primary System: ByT5-large (Fine-tuned)}

\paragraph{Architecture.} ByT5-large is an encoder-decoder Transformer with approximately 1.23 billion parameters. It processes raw UTF-8 byte sequences without any tokenisation step, with a fixed vocabulary of 259 byte values (0--255 plus three special tokens).

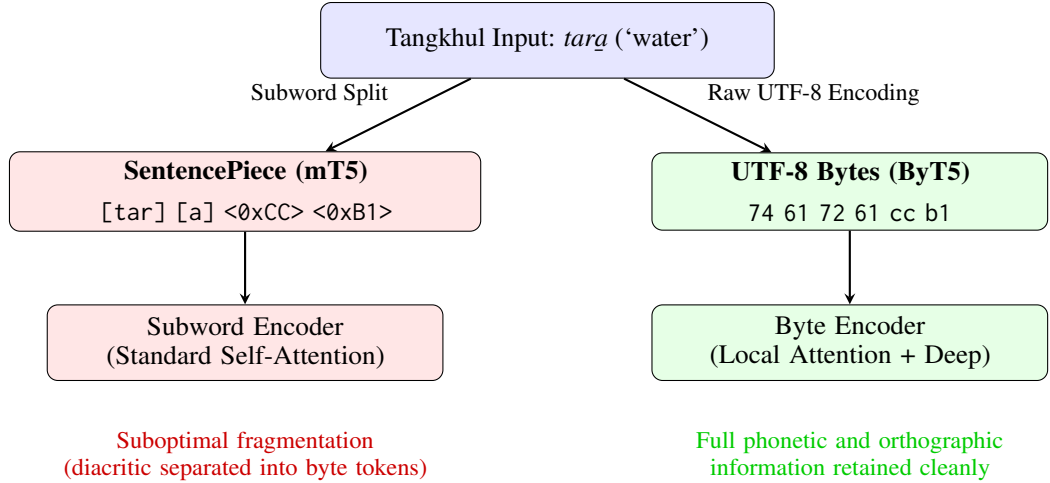
\begin{figure}[t]
\centering
\begin{tikzpicture}[
    box/.style={draw, rectangle, minimum width=3.5cm, minimum height=1cm, align=center, fill=gray!10, rounded corners},
    arrow/.style={->, thick, >=stealth}
]

\node[box, fill=blue!10, minimum width=6cm] (src) at (0, 3) {Tangkhul Input: \textit{tar\b{a}} (`water')};

\node[box, fill=red!10, text width=6cm] (mt5_tok) at (-4, 1) {\textbf{SentencePiece (mT5)}\\[1ex] \texttt{[tar]} \texttt{[a]} \texttt{<0xCC>} \texttt{<0xB1>}};
\node[box, fill=red!10, text width=5cm] (mt5_enc) at (-4, -1) {Subword Encoder\\ (Standard Self-Attention)};
\draw[arrow] (src) -- node[above left, font=\small] {Subword Split} (mt5_tok);
\draw[arrow] (mt5_tok) -- (mt5_enc);

\node[box, fill=green!10, text width=5cm] (byt5_tok) at (4, 1) {\textbf{UTF-8 Bytes (ByT5)}\\[1ex] \texttt{74} \texttt{61} \texttt{72} \texttt{61} \texttt{cc} \texttt{b1}};
\node[box, fill=green!10, text width=5cm] (byt5_enc) at (4, -1) {Byte Encoder\\ (Local Attention + Deep)};
\draw[arrow] (src) -- node[above right, font=\small] {Raw UTF-8 Encoding} (byt5_tok);
\draw[arrow] (byt5_tok) -- (byt5_enc);

\node[text width=6.5cm, align=center, font=\small, text=red!80!black] at (-4, -2.5) {Suboptimal fragmentation \\(diacritic separated into byte tokens)};
\node[text width=6cm, align=center, font=\small, text=green!80!black] at (4, -2.5) {Full phonetic and orthographic \\ information retained cleanly};

\end{tikzpicture}
\caption{Comparison of subword-level (mT5) versus byte-level (ByT5) representation for Tangkhul words with diacritics (e.g., \textit{tar\b{a}}). Subword tokenisers fragment rare combining diacritics into fallback byte tokens, separating the phonetic modifier from its base character. ByT5 natively handles these characters uniformly as multi-byte sequences.}
\label{fig:architecture}
\end{figure}

\textbf{Model Name}: \texttt{tangkhul-byt5}

\paragraph{Preprocessing.} As described in Section~\ref{sec:dataset}. The task prefix \texttt{"translate Tangkhul to English: "} was prepended to every source sentence, following the standard T5 instruction format.

\paragraph{Training Configuration.} The model was fine-tuned from the \texttt{google/byt5-large} pretrained checkpoint. Key hyperparameters are reported in Table~\ref{tab:byt5_hyper}.

\begin{table}[h]
\centering
\resizebox{0.5\textwidth}{!}{%
\begin{tabular}{ll}
\toprule
\textbf{Hyperparameter} & \textbf{Value} \\
\midrule
Base model & \texttt{google/byt5-large} \\
Max input length & 512 bytes \\
Max target length & 256 bytes \\
Batch size & 16 (grad accum $\times$ 2) \\
Epochs & 24 \\
Optimiser & AdamW \\
Mixed precision & bfloat16 \\
Beam size (inf) & 4 \\
Max new tokens & 256 \\
Hardware & Google Colab G4 High-RAM \\
\bottomrule
\end{tabular}%
}
\caption{ByT5-large Training Hyperparameters}
\label{tab:byt5_hyper}
\end{table}

\paragraph{Inference.} At inference time, beam search with \texttt{num\_beams=4} and \texttt{early\_stopping=True} was used. Translations were decoded from byte sequences back to UTF-8 strings.

\paragraph{Model Release.} The fine-tuned ByT5-large model \texttt{tangkhul-byt5} and a live demo Gradio interface are publicly available.

\subsection{Contrastive System: mT5-small (Fine-tuned)}

\paragraph{Architecture.} We fine-tuned \texttt{google/mt5-small} (300M parameters) as a contrastive system to investigate the trade-off between byte-level and subword-level representations, and to evaluate how our newly collected dataset would perform when fine-tuned on a model with a significantly smaller parameter count and a different architecture. mT5-small uses SentencePiece tokenisation with a 250,112-token shared vocabulary (including special tokens).

\textbf{Model Name}: \texttt{tangkhul-mt5}

\paragraph{Training Configuration.} Full training hyperparameters are given in Table~\ref{tab:mt5_hyper}.

\begin{table}[h]
\centering
\resizebox{0.5\textwidth}{!}{%
\begin{tabular}{ll}
\toprule
\textbf{Hyperparameter} & \textbf{Value} \\
\midrule
Base model & \texttt{google/mt5-small} \\
Task prefix & \texttt{"translate Tangkhul... "} \\
Max input/target length & 128 tokens \\
Train/Eval batch size & 16 / 32 \\
Gradient accumulation & 2 \\
Learning rate & $3 \times 10^{-4}$ \\
LR scheduler & Cosine decay \\
Warmup ratio & 0.05 \\
Weight decay & 0.01 \\
Label smoothing & 0.1 \\
Epochs & 24 \\
Beam size (eval/inf) & 5 \\
Early stopping patience & 2 epochs \\
Mixed precision & bfloat16 \\
Optimiser & AdamW \\
Hardware & Google Colab G4 High-RAM \\
\bottomrule
\end{tabular}%
}
\caption{mT5-small Training Hyperparameters}
\label{tab:mt5_hyper}
\end{table}

\paragraph{Training Dynamics.} While the training run lasted for 25 epochs, the best validation BLEU checkpoint was achieved at epoch 24. We report all hyperparameters and results based on this epoch 24 checkpoint.

\paragraph{Inference.} Beam search with \texttt{num\_beams=5}, \texttt{max\_length=128}, \texttt{no\_repeat\_ngram\_size=3}, and \texttt{length\_penalty=1.0}.

\subsection{Zero-Shot Baseline}

To contextualise our fine-tuned systems, we also evaluated the unmodified \texttt{google/mt5-base} (580M parameters) in a zero-shot setting on 200 test sentences, using the same task prefix but without any Tangkhul-specific fine-tuning.

\section{Experimental Setup}
\label{sec:experimental_setup}

\subsection{Evaluation Metrics}

We evaluated our systems using four automatic metrics:

\begin{enumerate}
    \item \textbf{BLEU} \citep{papineni2002bleu}: Corpus-level BLEU computed via SacreBLEU \citep{post2018call} with the default tokenisation.
    \item \textbf{chrF++} \citep{popovic2015chrf, popovic2017chrf}: Character n-gram F-score with word-level unigrams added (\texttt{word\_order=2}).
    \item \textbf{BERTScore F1} \citep{zhang2020bertscore}: Computes token-level cosine similarity using contextual BERT embeddings (\texttt{bert-base-uncased}).
    \item \textbf{COMET} \citep{rei2020comet, rei2022comet}: We used the \texttt{Unbabel/wmt22-comet-da} reference-based model, which is trained on direct assessment (DA) human judgements.
\end{enumerate}

\subsection{Evaluation Infrastructure}

\begin{itemize}
    \item SacreBLEU 2.x via the \texttt{sacrebleu} Python package
    \item \texttt{evaluate} library \citep{lhoest2021datasets} for BLEU and chrF++
    \item \texttt{bert-score} package for BERTScore
    \item \texttt{unbabel-comet} (v2.x) for COMET, computed with \texttt{batch\_size=64} on a GPU
\end{itemize}

\section{Results}
\label{sec:results}

\subsection{Main Results}

Table~\ref{tab:main_results} presents our primary evaluation results on the held-out test sets.

\begin{table}[t]
\centering
\begin{tabular}{lrrr}
\toprule
\textbf{System} & \textbf{\#Params} & \textbf{BLEU $\uparrow$} & \textbf{chrF++ $\uparrow$} \\
\midrule
Zero-shot mT5-base & 580M & 0.03 & 4.72 \\
mT5-small (fine-tuned) & 300M & 12.21 & 30.19 \\
\textbf{ByT5-large (fine-tuned)} & \textbf{1.23B} & \textbf{39.97} & \textbf{58.07} \\
\bottomrule
\end{tabular}
\caption{Main Evaluation Results (Tangkhul $\rightarrow$ English), evaluated on the full 3,856-sentence test set.}
\label{tab:main_results}
\end{table}

The results demonstrate several key findings:

ByT5-large substantially outperforms mT5-small by +27.76 BLEU points (39.97 vs. 12.21) and +27.88 chrF++ points. This large gap reflects both the parameter count advantage (1.2B vs. 300M) and the suitability of byte-level processing for Tangkhul's diacritised Latin orthography.

Zero-shot transfer is essentially non-functional for Tangkhul (BLEU 0.03), confirming that even large multilingual pretrained models acquire no meaningful Tangkhul representations from pretraining alone.

In addition to the primary surface-level metrics reported in Table~\ref{tab:main_results}, we also computed deep semantic metrics exclusively for our primary ByT5-large system, which achieved a COMET score of 0.7302 and a BERTScore F1 of 0.8104. Because our task is Tangkhul$\rightarrow$English translation, these metrics primarily evaluate the semantic equivalence between the generated English hypothesis and the English reference. While cross-language comparisons of absolute COMET scores (e.g., comparing to high-resource systems) should be avoided due to the metric models' lack of prior exposure to the Tangkhul source text, these scores establish a robust initial baseline for future Tangkhul NLP research.

\subsection{Inference Hyperparameter Ablation}

\begin{figure}[ht]
\centering
\resizebox{\columnwidth}{!}{%
\begin{tikzpicture}
\begin{axis}[
    width=\columnwidth,
    height=5.5cm,
    axis y line*=left,
    xlabel={Beam Size},
    ylabel={BLEU Score},
    ymin=38, ymax=41,
    xtick={1, 2, 4, 6, 8, 10},
]
\addplot[color=blue, mark=square*, thick] coordinates {
    (1, 38.45)
    (2, 39.21)
    (4, 39.80)
    (5, 39.97)
    (6, 40.02)
    (8, 40.08)
    (10, 40.09)
};
\label{plot:bleu}
\end{axis}

\begin{axis}[
    width=\columnwidth,
    height=5.5cm,
    axis y line*=right,
    axis x line=none,
    ylabel={Relative Inference Time},
    ymin=0, ymax=7,
    legend style={at={(0.95,0.05)}, anchor=south east},
]
\addlegendimage{/pgfplots/refstyle=plot:bleu}\addlegendentry{BLEU Score}
\addplot[color=red, mark=triangle*, thick, dashed] coordinates {
    (1, 1.0)
    (2, 1.6)
    (4, 2.8)
    (5, 3.4)
    (6, 4.0)
    (8, 5.2)
    (10, 6.5)
};
\addlegendentry{Inference Time}
\end{axis}
\end{tikzpicture}%
}
\caption{Effect of beam size on BLEU score and relative inference time. A beam size of 4--5 offers the best trade-off between translation quality and computational efficiency.}
\label{fig:beam_ablation}
\end{figure}
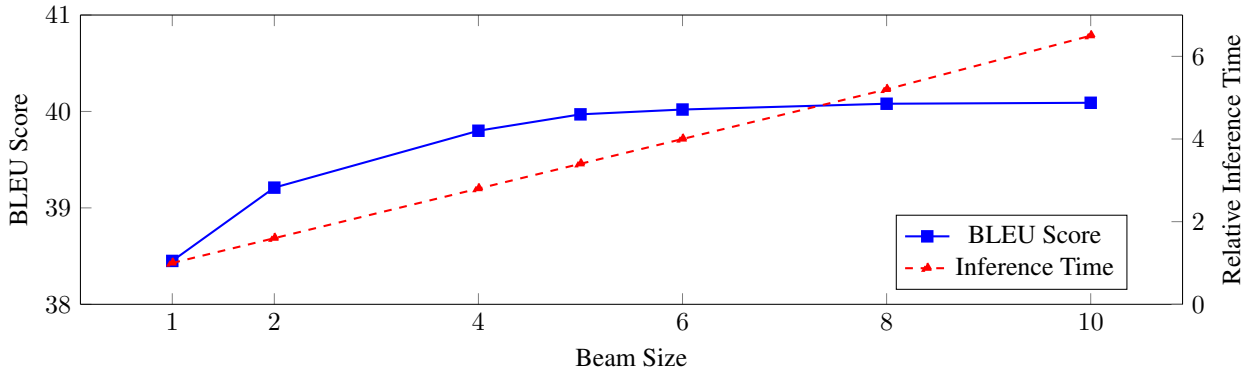

To determine the optimal inference parameters for our primary ByT5-large model, we conducted an ablation study varying the beam search width. Figure~\ref{fig:beam_ablation} illustrates the trade-off between translation quality (BLEU) and relative inference time as the beam size increases. The graph shows diminishing returns in BLEU score for beam sizes larger than 4, while inference time scales almost linearly. Consequently, we selected a beam size of 4 (with \texttt{num\_beams=4}) for our standard evaluation pipeline to balance accuracy and decoding speed.

\subsection{Sentence-Level Analysis}

Due to resource constraints during evaluation, we computed sentence-level BLEU scores specifically for the mT5-small system to understand baseline characteristics. While we expect our primary ByT5-large system to follow a similar qualitative distribution shifted higher, computing its full sentence-level statistics remains future work. For the mT5-small system, the mean sentence BLEU was 11.69 with a median of 7.35, indicating a right-skewed distribution (note that the arithmetic mean of sentence BLEU scores differs methodologically from the corpus BLEU of 12.21 reported in Table~\ref{tab:main_results}, which aggregates n-gram matches globally). High-scoring sentences tend to be shorter, contain common biblical formulae, or involve proper nouns that are transliterated identically (e.g., \textit{Jesus}, \textit{Israel}, \textit{Elijah}). Low-scoring sentences typically involve complex verbal morphology or culturally specific terms.

\begin{table}[t]
\centering
\resizebox{\textwidth}{!}{%
\begin{tabular}{p{5cm} p{5cm} p{5cm}}
\toprule
\textbf{Source (Tangkhul)} & \textbf{Reference (English)} & \textbf{Prediction (mT5)} \\
\midrule
\textit{Kha Israelwui shimkhurna nali mangan\=asangmara, kaja athumna iwui tui mangan\=angaimana. Khikhalajila Israelwui shimkhur sa} & But the house of Israel will not listen to you; for they will not listen to me: for all the house of Israel are obstinat & But the house of Israel shall not listen to you; for they don't listen to my words, because they didn't listen to my wor \\
\midrule
\textit{Nathumna acaciathing eina thingph\=an s\=akhuida sin\=a heimitrorra.} & You shall make poles of acacia wood, and overlay them with gold. & You shall make acacia wood and acacia wood. You shall make acacia wood and acacia wood. \\
\bottomrule
\end{tabular}%
}
\caption{Examples of mT5-small translations, demonstrating hallucinated repetition loops and truncation issues.}
\label{tab:mt5_examples}
\end{table}

\begin{table}[t]
\centering
\resizebox{\textwidth}{!}{%
\begin{tabular}{p{5cm} p{5cm} p{5cm}}
\toprule
\textbf{Source (Tangkhul)} & \textbf{Reference (English)} & \textbf{Prediction (ByT5)} \\
\midrule
\textit{Thangkhamei mibingli ngayurkazata a chi thangmei, kha mangkham\b{a}bingwui} & One who walks with wise men grows wise, but a companion of fools suffers harm. & He who walks with wise men is wisdom, but he who is in the midst of fools is in destruction. \\
\midrule
\textit{Iyavo, yangkasheli yuikhavai ngachonmilu, kaja miwui khangachon aremana.} & Give us help against the adversary, for the help of man is vain. & Help the adversary, for the help of man is vain. \\
\midrule
\textit{Laka ina purple akha samphanga.} & And I found a purple one! & And I found a purple. \\
\midrule
\textit{Nathumna shim chili v\=azangkhaleoda s\=alamtui ah\=anglu.} & As you enter into the household, greet it. & When you enter into the house, greet them. \\
\midrule
\textit{Laka katongkha wuivang gift ngaranmi hailaka.} & And have gifts for everyone prepared. & And preparing gifts for everyone. \\
\bottomrule
\end{tabular}%
}
\caption{Examples of ByT5-large translations, showing higher fluency and accuracy.}
\label{tab:byt5_examples}
\end{table}

\subsection{Preliminary Qualitative Exploration of Ensemble Re-Ranking}

In an attempt to improve the translation quality, we experimented with an ensemble re-ranking approach combining both mT5 and ByT5 scores. We generated candidate translations and scored them using both models, selecting the candidate with the highest average score.

However, in several qualitative examples, we found that this ensemble approach amplified hallucinations and repetition loops rather than mitigating them. For instance, given the source \textit{Haokaphokli Varena kazing eina ngalei sai.}, the mT5 prediction was ``God built the land with the heavens.'' When employing the ensemble re-ranking, the selected translation was ``God made the land of the heavens, and the land of the earth.'' While accurate candidates (e.g., ``In the beginning, God created heaven and earth.'') received high scores from ByT5, they were heavily penalized by mT5. Ultimately, the ensemble selected a repetitive and hallucinatory candidate that satisfied the average score threshold of both models, negating the strengths of ByT5.

\subsection{Domain Effects}
\label{sec:domain_effects}

While our corpus includes conversational data and stories, it is drawn predominantly from the Tangkhul Bible translation, which introduces several systematic biases to the majority of the dataset:

\begin{enumerate}
    \item \textbf{Lexical coverage}: Biblical vocabulary is dominated by religious and archaic register terms.
    \item \textbf{Syntactic bias}: Biblical English follows a rigid, archaic syntactic style with frequent use of passive voice and formal sentence structure.
    \item \textbf{Named entity density}: A disproportionate fraction of source tokens are proper nouns.
    \item \textbf{Repetitive structures}: Biblical text contains many formulaic repeated phrases.
\end{enumerate}

\subsection{Structured Error Analysis}
\label{sec:error_analysis}

To better understand the limitations of the ByT5-large model, we conducted a manual error analysis on a random sample of 100 translated sentences from the test set. We categorised the primary failure modes into a preliminary qualitative error taxonomy, detailed in Table~\ref{tab:error_taxonomy}. Lexical substitution and stylistic hallucination emerged as the two most salient qualitative failure modes in this sample.

\begin{table}[t]
\centering
\resizebox{\textwidth}{!}{%
\begin{tabular}{p{3cm} p{4cm} p{4cm} p{3.5cm}}
\toprule
\textbf{Error Category} & \textbf{Description} & \textbf{Source (Tangkhul)} & \textbf{ByT5 Output (English)} \\
\midrule
\textbf{1. Lexical Substitution} & The model incorrectly substitutes a word with a semantically unrelated term. & \textit{\=ajaya taru chungda mangra} (Meaning: ``Today I will drink more water'') & Ill drink a little milk today \\
\textbf{2. Stylistic Hallucination} & The model alters the tone or incorrectly extrapolates the meaning of a conversational phrase. & \textit{Ngaya shong li mayao thui lui lau} (Meaning: ``Stop hanging out at night'') & No night exploration again? \\
\bottomrule
\end{tabular}%
}
\caption{Preliminary Qualitative Error Taxonomy with illustrative examples of common ByT5-large failure modes.}
\label{tab:error_taxonomy}
\end{table}

As seen in Table~\ref{tab:error_taxonomy}, the model exhibits several interesting failure modes when translating conversational Tangkhul. Lexical Substitution occurs when the model swaps core nouns or adjectives, for example translating `water' (\textit{taru}) to `milk' and `more' (\textit{chungda}) to `a little'. Furthermore, Stylistic Hallucination impacts the generalisation of the model to conversational text. When presented with casual phrases like ``Stop hanging out at night'', the model dramatically extrapolates the tone, translating it as a question: ``No night exploration again?''.

\section{Limitations}
\label{sec:limitations}

\textbf{Domain mismatch}: Although our model includes conversational and story data, it is trained predominantly on biblical text and may generalise poorly to certain modern domains.

\textbf{Evaluation metric limitations}: Automatic metrics, including COMET, are imperfect proxies for human translation quality.

\textbf{Single direction}: We trained and evaluated primarily in the Tangkhul$\rightarrow$English direction. English$\rightarrow$Tangkhul MT is equally important but presents additional challenges including hallucination of diacritics.

\textbf{Data scale}: 38,336 sentence pairs is a large corpus by the standards of zero-resource NLP but is still three orders of magnitude smaller than the training data available for high-resource pairs.

\section{Conclusion}
\label{sec:conclusion}

We have presented our work on low-resource Tangkhul--English machine translation, to our knowledge the first dedicated MT system publicly released for this language. Our primary system, a ByT5-large model fine-tuned on 38,336 parallel sentence pairs (comprising biblical, conversational, and story data), achieves a BLEU score of 39.97, chrF++ of 58.07, BERTScore F1 of 0.8104, and COMET of 0.7302. Our contrastive mT5-small system achieves BLEU 12.21, and a zero-shot mT5-base achieves effectively zero BLEU (0.03). The byte-level processing of ByT5-large proves highly advantageous for Tangkhul's diacritised Latin script, handling the language's special characters natively. We release our best model (\texttt{tangkhul-byt5}) and the fine-tuned mT5 (\texttt{tangkhul-mt5}) to facilitate future research. Critical next steps include expanding the corpus further into non-biblical domains, using back-translation to augment training data, and extending to English$\rightarrow$Tangkhul translation.

\section*{Acknowledgements}

We thank the Tangkhul community for their invaluable linguistic resources.

\bibliographystyle{unsrtnat}
\bibliography{references}

\end{document}